\documentclass[conference,compsoc]{IEEEtran}
\IEEEoverridecommandlockouts

\usepackage[T1]{fontenc}
\usepackage{cite}
\usepackage{amsmath,amssymb,amsfonts}
\usepackage{graphicx}
\usepackage{booktabs}
\usepackage{textcomp}
\usepackage{url}

\renewcommand{\figurename}{Fig.}   
\graphicspath{{figs/}}

\newcommand{\statAbBaseCyber}{34.8\%}

\newcommand{\statAbCritCICyber}{[-6.3, -3.1]}

\newcommand{\statAbShufLetterCyber}{31.2\%}

\newcommand{\statAbShufPosCyber}{30.5\%}

\newcommand{\statAbStateCICyber}{[-3.3, -0.8]}

\newcommand{\statAbSymAccCICyber}{[-2.4, +0.5]}

\newcommand{\statAbSymAccDeltaCyber}{-1.0}

\newcommand{\statAbSymCyber}{36.1\%}
\newcommand{\statAccBio}{85.2\%}
\newcommand{\statAccBioRobust}{80.1\%}
\newcommand{\statAccCyber}{63.4\%}

\newcommand{\statAurocMax}{0.86}
\newcommand{\statAurocMin}{0.62}
\newcommand{\statAurocPooled}{0.820}
\newcommand{\statAurocPooledHi}{0.830}
\newcommand{\statAurocPooledLo}{0.810}
\newcommand{\statAurocSpread}{0.006}

\newcommand{\statBrierPmax}{0.161}
\newcommand{\statCascadeAdvantageBio}{+0.3}

\newcommand{\statCascadeCostSaving}{45\%}
\newcommand{\statCascadeCyberAtForty}{67.0\%}
\newcommand{\statCascadeCyberAtTwenty}{65.9\%}
\newcommand{\statCascadeCyberCallsForty}{2.2}
\newcommand{\statCascadeCyberRandForty}{64.8\%}
\newcommand{\statCascadeFracOfGain}{97\%}
\newcommand{\statCascadeRandBeatBio}{6.8\%}

\newcommand{\statChanceSupp}{21.3\%}
\newcommand{\statCollectionDates}{\mbox{2026-09-19} to \mbox{2026-09-20}}
\newcommand{\statConfAltEntropy}{20.7\%}
\newcommand{\statConfAltMargin}{31.0\%}
\newcommand{\statConfAltPmax}{28.0\%}
\newcommand{\statConfExample}{0.33}
\newcommand{\statConfMatch}{97.9\%}
\newcommand{\statConfTol}{0.015}
\newcommand{\statConsistBio}{89.4\%}
\newcommand{\statConsistCyber}{62.6\%}
\newcommand{\statDistinctProbs}{101}
\newcommand{\statEceConf}{0.091}
\newcommand{\statEceLitQA}{0.168}
\newcommand{\statEceMax}{0.193}
\newcommand{\statEceMin}{0.033}
\newcommand{\statEcePmax}{0.034}
\newcommand{\statEcePmaxHi}{0.045}
\newcommand{\statEcePmaxLo}{0.027}

\newcommand{\statEceSupp}{0.193}
\newcommand{\statEceWorstWmdp}{0.057}
\newcommand{\statEnsAccCyber}{67.1\%}
\newcommand{\statEnsGainBio}{+0.5}
\newcommand{\statEnsGainBioHi}{+1.4}
\newcommand{\statEnsGainBioLo}{-0.2}
\newcommand{\statEnsGainCyber}{+3.8}
\newcommand{\statEnsGainCyberHi}{+5.3}
\newcommand{\statEnsGainCyberLo}{+2.4}
\newcommand{\statErrGivenHiBio}{3.9\%}
\newcommand{\statErrGivenHiCyber}{7.5\%}
\newcommand{\statErrGivenHiLitQA}{38.1\%}
\newcommand{\statErrGivenHiLitQACI}{[21,\,59]}

\newcommand{\statErrRatioBio}{4.64}
\newcommand{\statErrRatioCyber}{2.93}
\newcommand{\statErrStableCyber}{21.3\%}
\newcommand{\statErrUnstableCyber}{62.5\%}

\newcommand{\statFisherBioCyberP}{0.00469}
\newcommand{\statFisherBioCyberQ}{0.0141}

\newcommand{\statInconsistCyber}{37.4\%}
\newcommand{\statInputTokens}{13.5}
\newcommand{\statLabMax}{61.8\%}

\newcommand{\statLabMin}{36.6\%}

\newcommand{\statLlmCheapRatio}{1.3}
\newcommand{\statLlmCheapUsd}{0.71}
\newcommand{\statLlmMidRatio}{148}
\newcommand{\statLlmMidUsd}{84}
\newcommand{\statLlmTopRatio}{742}
\newcommand{\statLlmTopUsd}{421}
\newcommand{\statMajAccCyber}{65.2\%}
\newcommand{\statModelVersion}{jev-1.13.0}
\newcommand{\statNBootCI}{2{,}000}

\newcommand{\statNCallsCircular}{13{,}040}

\newcommand{\statNCallsTotal}{19{,}060}
\newcommand{\statNCircItemsBio}{1{,}273}
\newcommand{\statNCircItemsCyber}{1{,}987}

\newcommand{\statNDatasets}{10}

\newcommand{\statNErrors}{1}
\newcommand{\statNHiBio}{848}
\newcommand{\statNHiCyber}{550}
\newcommand{\statNHiLitQA}{21}

\newcommand{\statNItems}{6{,}020}
\newcommand{\statNItemsRaw}{6{,}021}
\newcommand{\statNSigPosDatasets}{3}
\newcommand{\statNTestedPosDatasets}{8}

\newcommand{\statNullRatioBio}{10.03}

\newcommand{\statNullRatioCyber}{5.43}
\newcommand{\statNullRatioCyberCI}{[4.68, 6.35]}

\newcommand{\statPosGoldACyber}{26.6\%}

\newcommand{\statPosHighP}{0.84}
\newcommand{\statPosHighPower}{1.00}

\newcommand{\statPosLowGoldA}{21.8\%}
\newcommand{\statPosLowP}{<10^{-4}}
\newcommand{\statPosLowPredA}{42.5\%}

\newcommand{\statPosPBio}{0.11}
\newcommand{\statPosPCyber}{<10^{-4}}
\newcommand{\statPosPowerBio}{1.00}
\newcommand{\statPosPredACyber}{34.8\%}

\newcommand{\statRCBioAtSixtyPct}{97.1\%}

\newcommand{\statRCCyberAtSixtyPct}{78.9\%}
\newcommand{\statRCCyberBase}{0.634}

\newcommand{\statReBetweenCyber}{37.8\%}
\newcommand{\statReCalls}{25{,}040}
\newcommand{\statReFacItems}{500}
\newcommand{\statReFacRepeats}{3}

\newcommand{\statReGainDiffCICyber}{[+2.0, +4.8]}
\newcommand{\statReGainDiffCyber}{+3.4}

\newcommand{\statReGainIdentCyber}{+0.5}

\newcommand{\statReGainRotCyber}{+3.8}

\newcommand{\statReGapCyber}{28.2}
\newcommand{\statReIdentBio}{2.4\%}
\newcommand{\statReIdentCyber}{9.2\%}
\newcommand{\statReNoiseShareBio}{23\%}
\newcommand{\statReNoiseShareCyber}{25\%}
\newcommand{\statReRatioBio}{3.8}
\newcommand{\statReRatioCyber}{4.8}
\newcommand{\statReRepeats}{4}
\newcommand{\statReRotBio}{10.6\%}
\newcommand{\statReRotCyber}{37.4\%}

\newcommand{\statReWithinCyber}{7.8\%}
\newcommand{\statRotSpreadBio}{1.3}
\newcommand{\statRotSpreadCyber}{0.8}
\newcommand{\statSingleAccCyber}{63.3\%}
\newcommand{\statTieContainingFirst}{46}
\newcommand{\statTieExpectedFirst}{23.0}
\newcommand{\statTieFrac}{1.2\%}
\newcommand{\statTieObservedFirst}{21}
\newcommand{\statTieTotal}{71}
\newcommand{\statTokensPerCall}{709}
\newcommand{\statUsdPerM}{0.042}
\newcommand{\statUsdTotal}{0.57}

\newcommand{\jev}{\textsc{Jev}}
\newcommand{\pmax}{p_{\max}}

\makeatletter
\def\IEEEkeywords{\normalfont\@IEEEtweakunitybaselinestretch{1.15}\bfseries
    \if@twocolumn
      \@IEEEabskeysecsize\vskip 0.5\baselineskip plus 0.25\baselineskip minus 0.25\baselineskip\noindent
      \textit{\IEEEkeywordsname:}\ \relax
    \else
      \bgroup\par\addvspace{0.5\baselineskip}\centering\@IEEEabskeysecsize\textbf{\IEEEkeywordsname}\par\addvspace{0.5\baselineskip}\egroup\quotation\@IEEEabskeysecsize%
    \fi\@IEEEgobbleleadPARNLSP}
\makeatother

\begin{document}

\title{Auditing System-1 Models on Biosecurity-Relevant Benchmarks:\\
Calibration, Selective Prediction, and Permutation Instability\\
in a Non-Generative Model}

\author{%
\IEEEauthorblockN{Kimon Antonios Provatas \qquad
                  Ilias Georgakopoulos-Soares}
\thanks{Corresponding author: Ilias Georgakopoulos-Soares.}
\IEEEauthorblockA{Division of Pharmacology and Toxicology, College of Pharmacy\\
The University of Texas at Austin, Dell Pediatric Research Institute\\
Austin, TX, USA\\
kap4722@my.utexas.edu \qquad ilias@austin.utexas.edu}
}

\maketitle

\begin{abstract}
Non-generative ``System-1'' models return structured probabilistic decisions in a single
forward pass, without autoregressive decoding, at a small fraction of the inference cost of
a generative model. This makes them of interest as inexpensive components in larger
pipelines, but their reliability on biosecurity-relevant tasks has not been systematically
examined. We audit one commercial System-1 model on \statNItems{} multiple-choice items
drawn from the Weapons of Mass Destruction Proxy (WMDP), a paraphrase-robust WMDP-Bio
variant, and six LAB-Bench subtasks, measuring accuracy, calibration, error detection,
selective prediction, and sensitivity to the order in which answer options are presented.
Accuracy is strongly task-dependent. Once the vendor's uncertainty field is correctly
interpreted, the model is reasonably well calibrated (pooled expected calibration error
\statEcePmax{}) and its top-1 probability separates correct from incorrect predictions
(pooled AUROC \statAurocPooled{}), though both degrade substantially on the weaker tasks.
Under four cyclic rotations of the answer options, \statInconsistCyber{} of WMDP-Cyber
items receive different answers; a control using byte-identical repeated calls attributes
most of this to option order rather than run-to-run variation. Averaging probabilities
across rotations improves WMDP-Cyber accuracy by \statEnsGainCyber{}~percentage points,
and applying it only to low-confidence items recovers most of that gain at well under the
cost of averaging every item.
\end{abstract}


\section{Introduction}

Recent work on efficient inference has produced models that return a typed, probabilistic
decision from a single forward pass rather than by decoding tokens autoregressively. Such
``System-1'' models are offered for settings in which many decisions must be made cheaply.
The evaluation reported here cost \statUsdTotal{}~USD for \statNCallsTotal{} calls.
Answering the same items with a generative model at current list prices would cost
between roughly \statLlmCheapUsd{}~USD, using the cheapest small model and emitting only
the option label, and \statLlmTopUsd{}~USD at frontier prices with a short chain of
thought (Section~\ref{sec:setup}). That difference is what makes repeated evaluation of
the same item a practical experimental option here rather than a hypothetical one.

Low-cost probabilistic models are potentially relevant to AI safety and biosecurity
pipelines, where auxiliary classifiers are already deployed alongside larger
models~\cite{llamaguard,constclass}. Before such a role could be considered, the basic
empirical properties of these models need to be established: how accurate they are, how
well their reported uncertainty tracks correctness, whether that uncertainty supports
abstention, and how stable their decisions are under changes that do not alter the meaning
of the input. Benchmark accuracy addresses only the first of these.

This paper reports a reliability audit of one commercially available System-1 model,
\jev{}, on two kinds of benchmark. The Weapons of Mass Destruction Proxy (WMDP)~\cite{wmdp} is a multiple-choice set
covering biosecurity, chemical security, and cybersecurity, written as a measurable
stand-in for knowledge that could assist misuse and introduced so that unlearning
methods could be scored on \emph{reducing} it; high accuracy on it is therefore not a
capability result. LAB-Bench~\cite{labbench} covers practical biology research skill and
carries the opposite sign. Neither benchmark asks the model to judge whether a request is dangerous,
so neither measures screening or hazard-detection performance. They are used here as
subject matter on which reliability, uncertainty, and robustness can be studied.

We ask four questions. When is the model correct? Does its reported uncertainty identify
likely errors? Can uncertain items be set aside to raise accuracy on the remainder? And are
its decisions stable under a change that does not affect meaning, namely the order in which
the answer options are listed?

\noindent This paper makes three contributions.
\begin{enumerate}
\item A reliability audit of a commercial non-generative System-1 model across
      \statNItems{} biosecurity- and biology-relevant benchmark items, covering accuracy,
      calibration, error detection, and selective prediction.
\item A controlled answer-order study, with a byte-identical-repeat control that
separates option-order sensitivity from run-to-run variation, showing substantial item-level instability despite comparatively stable aggregate accuracy.
\item An evaluation of selective permutation averaging, showing that repeated inference
      applied only to low-confidence items improves accuracy on the instability-heavy task
      while avoiding the cost of evaluating every item four times.
\end{enumerate}

We do not claim novelty for confidence-based selection of items for further
computation~\cite{calm,bild,frugalgpt,routellm}, nor for the observation that
multiple-choice evaluation is sensitive to option order~\cite{alzahrani,gupta}.

\section{Background and Related Work}

\textbf{Decision cascades and early exit.} Spending extra computation only on
inputs a cheap model finds difficult is established: CALM~\cite{calm} exits a decoder
early under a calibrated criterion, Big Little Decoder~\cite{bild} escalates when a
small model's maximum predicted probability falls below a threshold, and
FrugalGPT~\cite{frugalgpt} and RouteLLM~\cite{routellm} allocate queries across models
under a cost budget. Our final experiment applies the same idea with additional answer
permutations of the same model rather than a larger model. For uncertainty we use
standard tools: maximum class probability as an error-detection
baseline~\cite{hendrycks}, expected calibration error and reliability
diagrams~\cite{guo}, and risk--coverage analysis from selective
classification~\cite{geifman}.

\textbf{Multiple-choice robustness.} Reordering answer options changes model rankings on leaderboards~\cite{alzahrani} and
degrades accuracy in model-dependent ways~\cite{gupta}; Zheng et al.~\cite{zheng}
attribute the effect to a token-level bias toward particular option labels, and
formatting changes that preserve meaning can also shift accuracy
substantially~\cite{formatspread}. That work concerns generative models scored by likelihood
over option tokens. We test whether a non-generative model emitting a typed distribution in
one pass shows comparable sensitivity.

\textbf{Safety classifiers and biosecurity context.} Deployed language-model systems commonly place auxiliary classifiers around a primary
model to detect disallowed content: moderation classifiers~\cite{openaimod}, input--output safeguards such as Llama Guard~\cite{llamaguard} and
ShieldGemma~\cite{shieldgemma}, constitutional classifiers trained to resist
jailbreaks~\cite{constclass}, with biological risk a specific concern~\cite{biorisk}.
That work motivates why inexpensive probabilistic models are of interest, but we
evaluate none of these safeguard tasks: prior work asks whether a system can block a
dangerous request, whereas we ask what reliability properties a cheap System-1 model
exhibits on biosecurity-relevant knowledge benchmarks.

\section{Experimental Setup}
\label{sec:setup}

\jev{} is a commercial non-generative model queried through the
\texttt{v1} REST endpoint. A request carries a free-text state and a typed question; the
response gives a selected category, a probability distribution over categories, a scalar
\texttt{confidence}, the resolved model version, and token usage. All calls used the
alias \texttt{jev-latest}, which resolved to \texttt{\statModelVersion{}} throughout,
and were made from \statCollectionDates{}. The model is proprietary: no logits, token
probabilities, or random seed are exposed, and returned probabilities are quantised to
two decimals (\statDistinctProbs{} distinct values observed). Because inference is a
single pass rather than autoregressive decoding, per-item cost is low; we did not
measure latency and make no claim about it.

Table~\ref{tab:main} lists the \statNDatasets{} datasets. WMDP-Bio,
WMDP-Chem, and WMDP-Cyber are hazardous-knowledge proxies~\cite{wmdp}, WMDP-Bio-Robust is a
paraphrased variant of WMDP-Bio, and six LAB-Bench subtasks cover practical biology
research~\cite{labbench}. Option counts range from 2 to 10, so we report accuracy against
each item's own chance level $1/n$ rather than a single 25\% line. LAB-Bench items are
formed by combining the reference answer with its distractors under a fixed shuffle seed;
items with fewer than two distinct options after de-duplication are skipped. Of \statNItemsRaw{} single-pass calls, \statNErrors{} returned an API error and is
excluded, leaving \statNItems{} items.

We report accuracy with Wilson intervals; expected calibration error
(ECE) over ten equal-width bins; the area under the receiver operating characteristic curve (AUROC) for separating
correct from incorrect predictions using $\pmax$, the largest returned probability; the error rate conditional on high
confidence, $P(\text{error} \mid \pmax \geq 0.9)$; and risk--coverage curves obtained by
withholding the least confident items. Intervals on proportions are Wilson; intervals on
ECE, AUROC, and on differences between accuracies are percentile bootstrap over items,
\statNBootCI{} resamples with a fixed seed.

For the two four-option WMDP suites we additionally present each
item under all four cyclic rotations of its option list ($\statNCircItemsBio{}$ and
$\statNCircItemsCyber{}$ items, \statNCallsCircular{} calls), mapping each returned
distribution back to the original option indices so that rotations are comparable.

The endpoint reports token usage per response. The study consumed
\statInputTokens{}~million input tokens over \statNCallsTotal{} calls, a mean of
\statTokensPerCall{} per call; at \statUsdPerM{}~USD per million input tokens with no
output charge, \statUsdTotal{}~USD. The same calls with a generative model, at list
prices retrieved 2026-09-20, would cost about \statLlmCheapUsd{}~USD for the cheapest
small model emitting only an option label (\statLlmCheapRatio{}$\times$),
\statLlmMidUsd{}~USD mid-tier with a short chain of thought
(\statLlmMidRatio{}$\times$), and \statLlmTopUsd{}~USD at frontier prices
(\statLlmTopRatio{}$\times$). These are upper bounds: batch interfaces list at half rate, cached input at a tenth, and
large deployments negotiate below list, while an agentic configuration issuing several
turns per decision moves the other way. We did not measure latency.

\begin{table*}[t]
\centering
\caption{Per-dataset results. \emph{Lift} is chance-adjusted accuracy, $(\mathrm{acc}-c)/(1-c)$, using each item's own option count $c=1/n$ rather than a single $25\%$ line, since option counts range from 2 to 10. ECE and AUROC are computed on $p_{\max}$ rather than on the vendor \texttt{confidence} field (Section~\ref{sec:uncertainty}). $P(\text{err}\mid p_{\max}\!\geq\!0.9)$ is the error rate given a confident answer, with Wilson intervals. \emph{Retained} is accuracy on the items kept when the least confident 40\% are withheld (Section~\ref{sec:selective}).}
\label{tab:main}
\footnotesize
\setlength{\tabcolsep}{3.3pt}
\begin{tabular}{@{}llrccccccc@{}}
\toprule
Suite & Dataset & $N$ & opts. & accuracy [95\% CI] & lift & ECE & AUROC & $P(\text{err}\mid p_{\max}\!\geq\!0.9)$ & retained \\
\midrule
WMDP & WMDP-Bio & 1273 & 4 & 0.852 [0.83, 0.87] & 0.802 & 0.036 & 0.855 & 0.039 [0.03, 0.05] & 0.971 \\
 & WMDP-Bio-Robust & 811 & 4 & 0.801 [0.77, 0.83] & 0.735 & 0.044 & 0.823 & 0.063 [0.04, 0.09] & 0.938 \\
 & WMDP-Chem & 408 & 4 & 0.730 [0.69, 0.77] & 0.641 & 0.057 & 0.824 & 0.056 [0.03, 0.10] & 0.898 \\
 & WMDP-Cyber & 1987 & 4 & 0.634 [0.61, 0.66] & 0.512 & 0.033 & 0.774 & 0.075 [0.06, 0.10] & 0.789 \\
\midrule
LAB-Bench & LB-CloningScenarios & 32 & 4--8 & 0.469 [0.31, 0.64] & 0.314 & 0.126 & 0.671 & 0.500 [0.09, 0.91] & 0.579 \\
 & LB-ProtocolQA & 108 & 4--7 & 0.537 [0.44, 0.63] & 0.397 & 0.124 & 0.620 & 0.077 [0.01, 0.33] & 0.600 \\
 & LB-SeqQA & 600 & 2--5 & 0.618 [0.58, 0.66] & 0.492 & 0.093 & 0.861 & 0.000 [0.00, 0.05] & 0.825 \\
 & LB-DbQA & 520 & 4 & 0.398 [0.36, 0.44] & 0.197 & 0.086 & 0.653 & 0.091 [0.03, 0.28] & 0.468 \\
 & LB-LitQA2 & 199 & 2--10 & 0.402 [0.34, 0.47] & 0.208 & 0.168 & 0.656 & 0.381 [0.21, 0.59] & 0.479 \\
 & LB-SuppQA & 82 & 4--8 & 0.366 [0.27, 0.47] & 0.195 & 0.193 & 0.709 & 0.250 [0.07, 0.59] & 0.429 \\
\midrule
\multicolumn{2}{@{}l}{\textbf{pooled}} & 6020 & 2--10 & 0.673 [0.661, 0.685] & -- & 0.034 & 0.820 & -- & -- \\
\bottomrule
\end{tabular}
\end{table*}

\section{Results}

\subsection{Benchmark performance and uncertainty}
\label{sec:uncertainty}

Accuracy varies widely across tasks (Table~\ref{tab:main}). It is highest on WMDP-Bio
(\statAccBio{}), falls on the paraphrased WMDP-Bio-Robust (\statAccBioRobust{}), and is
substantially lower on WMDP-Cyber (\statAccCyber{}). LAB-Bench results are lower and more variable, from \statLabMin{} on SuppQA to
\statLabMax{} on SeqQA; SuppQA is the weakest subtask, though still above its
\statChanceSupp{} chance level. Aggregate accuracy therefore describes the
particular task at least as much as the model.

Interpreting the model's uncertainty requires first establishing what the reported
\texttt{confidence} field measures. Across all \statNItems{} records it matches
\[
c \;=\; \frac{\pmax - 1/n}{1 - 1/n}
\]
to within $\pm\statConfTol{}$ for \statConfMatch{} of records, where $n$ is the number of
options. Competing readings fit less well: raw $\pmax$ matches \statConfAltPmax{}, the
margin between the top two options \statConfAltMargin{}, and normalised entropy
\statConfAltEntropy{}. The field is a rescaling of the top-1 probability onto $[0,1]$,
mapping a uniform distribution to 0 and a one-hot distribution to 1; on a four-option item
an even split between two candidates ($\pmax = 0.5$) is reported as \statConfExample{}. It
is not a predicted probability of correctness, and calibration should be assessed against
$\pmax$ instead.

\begin{figure*}[t]
\centering
\includegraphics[width=\textwidth]{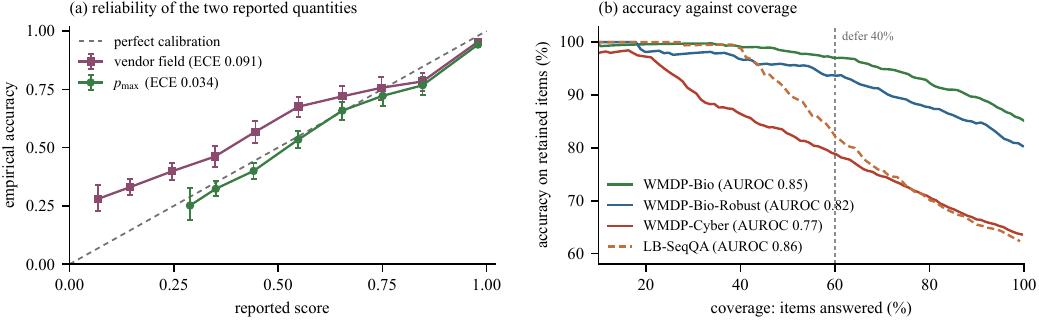}
\caption{Uncertainty and selective prediction. (a) Pooled reliability curves for the two
quantities the API reports: the \texttt{confidence} field (squares) is an affine
rescaling of $\pmax$ (circles), so calibration measured against it differs from
calibration measured against the probability itself. Bins holding fewer than 25 items
are omitted, because $\pmax \geq 1/n$ leaves the lowest bins of the $\pmax$ curve nearly
empty; error bars are Wilson intervals. (b) Accuracy on retained items against coverage,
withholding the least confident first, for the three WMDP suites and the largest LAB-Bench subtask (dashed); the remaining six subtasks are omitted for legibility and appear
in Table~\ref{tab:main}. The dashed vertical marker is the 40\% deferral operating point
quoted in Section~\ref{sec:selective} and in Table~\ref{tab:main}.}
\label{fig:uncertainty}
\end{figure*}

On that basis, pooled ECE is \statEcePmax{} (95\% CI $[\statEcePmaxLo{},
\statEcePmaxHi{}]$; Brier \statBrierPmax{}), against \statEceConf{} when the rescaled
field is used (Fig.~\ref{fig:uncertainty}(a)). Pooled AUROC for separating correct from
incorrect predictions is \statAurocPooled{} (95\% CI $[\statAurocPooledLo{},
\statAurocPooledHi{}]$), so $\pmax$ carries information about correctness. Both pooled figures flatter the model, since pooling lets $\pmax$ discriminate partly on
which dataset an item came from: per-task values span \statAurocMin{}--\statAurocMax{}
AUROC and \statEceMin{}--\statEceMax{} ECE (Table~\ref{tab:main}), and those are the
relevant figures for a system consuming these predictions. We compared four candidate signals, $\pmax$, the top-two margin, negative
entropy, and the rescaled field; they differ by at most \statAurocSpread{} AUROC, so the
choice among them is not consequential.

Calibration quality is not uniform. Per-dataset ECE is at most \statEceWorstWmdp{} on the
four WMDP suites but reaches \statEceLitQA{} on LitQA2 and \statEceSupp{} on SuppQA, where
item counts are small.

\subsection{Confidence and selective prediction}
\label{sec:selective}

A quantity that is directly interpretable for a system consuming these predictions is the
error rate given a confident answer. Under $P(\text{error} \mid \pmax \geq 0.9)$ the model is
most reliable on WMDP-Bio (\statErrGivenHiBio{}, $n=\statNHiBio{}$) and less so on WMDP-Cyber
(\statErrGivenHiCyber{}, $n=\statNHiCyber{}$); a Fisher exact test gives
$p = \statFisherBioCyberP{}$, and $q = \statFisherBioCyberQ{}$ after Benjamini--Hochberg
correction across all pairwise comparisons. Confident answers are considerably less reliable
on the LAB-Bench literature tasks, reaching \statErrGivenHiLitQA{} on LitQA2 (95\% CI
$\statErrGivenHiLitQACI{}\%$, $n=\statNHiLitQA{}$), although that subset is small. A
confidence threshold calibrated on one dataset should not be assumed to transfer to another.

Withholding low-confidence items raises accuracy on the remainder
(Fig.~\ref{fig:uncertainty}(b)). Deferring the least confident 40\% of items raises accuracy on the retained set from
\statAccBio{} to \statRCBioAtSixtyPct{} on WMDP-Bio and from \statAccCyber{} to
\statRCCyberAtSixtyPct{} on WMDP-Cyber. Confidence therefore identifies a
subset on which the model performs substantially better, which is a prerequisite for any
scheme that would direct difficult items elsewhere.

\begin{figure*}[!t]
\centering
\includegraphics[width=\textwidth]{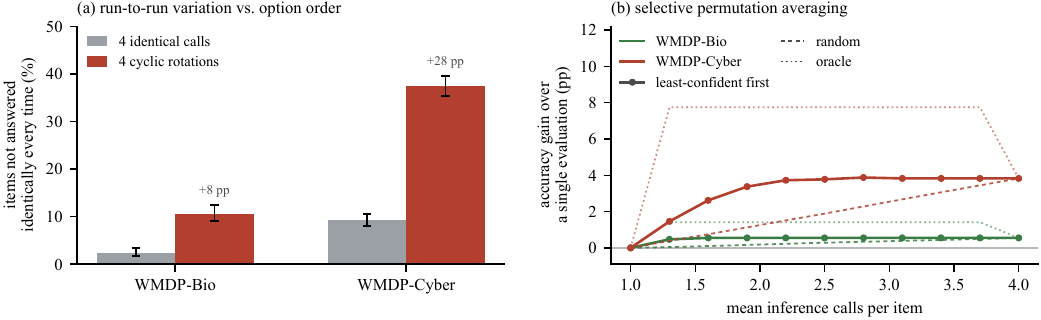}
\caption{Answer-order instability and selective permutation averaging. (a) Items
answered differently across four byte-identical calls, which isolates run-to-run
variation, against four cyclic rotations, which adds option order; the annotation gives
the difference. Wilson intervals. (b) Accuracy gain over a single evaluation against
mean inference calls per item, plotted as a gain so that two datasets 22~pp apart in
absolute accuracy share one scale. Items are selected for additional permutations least-confident first, and compared against random selection at the same budget (averaged over
400 draws) and against an oracle that selects, under a fixed budget, the items where
averaging helps.}
\label{fig:permutation}
\end{figure*}

\subsection{Answer-order instability}

We next ask whether decisions are stable under a change that does not alter the content of
the question. Each item is presented under all four cyclic rotations of its option list, and
each returned distribution is mapped back to the original option indices.

Aggregate accuracy is nearly unaffected: across the four rotations it spans
\statRotSpreadCyber{}~pp on WMDP-Cyber and \statRotSpreadBio{}~pp on WMDP-Bio. Item-level
agreement is much lower. Only \statConsistCyber{} of WMDP-Cyber items receive the same answer
under all four rotations, against \statConsistBio{} on WMDP-Bio
(Fig.~\ref{fig:permutation}(a)). Error rates differ sharply between the two groups: on
WMDP-Cyber, \statErrUnstableCyber{} of unstable items are answered incorrectly under the
original ordering against \statErrStableCyber{} of stable items. A substantial fraction of errors occurs on items whose predictions are unstable under
reordering.

One observation qualifies this. An association between instability and error is partly
implied by the measurement: consistency and correctness are both defined relative to the
first rotation, and there is one way to be consistently correct but several ways to be
wrong. A null model with no positional preference, matched to the same accuracy and the same
consistency, produces an error ratio of \statNullRatioCyber{} on WMDP-Cyber (95\% CI
$\statNullRatioCyberCI{}$) against the \statErrRatioCyber{} observed, and
\statNullRatioBio{} against \statErrRatioBio{} on WMDP-Bio. The observed ratio falls
\emph{below} what the artefact alone predicts. The descriptive result stands; a causal
reading does not.

\subsection{Separating option order from run-to-run variation}

Rotating the options changes
the presentation and also makes a fresh call, and the endpoint is not deterministic, so
the second alone could produce disagreement. We therefore held the presentation fixed
and queried every item \statReRepeats{} times with a byte-identical prompt, and queried
a stratified subsample of \statReFacItems{} items per dataset \statReFacRepeats{} times
at every rotation for a two-factor design (\statReCalls{} calls). Dispatch order was
shuffled so repeats of one item were never consecutive.
Identical prompts disagree far less often than rotations do. On WMDP-Cyber,
\statReIdentCyber{} of items receive different answers across \statReRepeats{} identical
calls, against \statReRotCyber{} across four rotations, leaving \statReGapCyber{}~pp
attributable to option order; on WMDP-Bio the figures are \statReIdentBio{} and
\statReRotBio{} (Fig.~\ref{fig:permutation}(a)). Run-to-run variation thus accounts for
roughly \statReNoiseShareCyber{} of the observed instability on WMDP-Cyber and
\statReNoiseShareBio{} on WMDP-Bio. The two-factor subsample agrees: within-rotation
disagreement is \statReWithinCyber{} against \statReBetweenCyber{} between rotations on
WMDP-Cyber, a ratio of \statReRatioCyber{}$\times$ (\statReRatioBio{}$\times$ on WMDP-Bio). The same design gives the matched-budget control for Section~\ref{sec:permavg}:
four identical calls cost exactly what four rotations cost. Averaging four identical
calls changes WMDP-Cyber accuracy by \statReGainIdentCyber{}~pp, with an interval
spanning zero, whereas averaging four rotations gives \statReGainRotCyber{}~pp; the
difference is \statReGainDiffCyber{}~pp (95\% CI $\statReGainDiffCICyber{}$). The gain
therefore comes from averaging over presentations, not from averaging away per-call
noise. Positional preference is directly detectable too. Since predicted labels are drawn
toward correct labels by accuracy, a $\chi^2$ against the correct-label distribution
loses power as accuracy rises, so we compare against a simulated null of the same
accuracy with no positional preference. The first option is over-selected on \statNSigPosDatasets{} of \statNTestedPosDatasets{}
testable datasets --- WMDP-Cyber, LB-SeqQA and LB-DbQA --- most clearly on WMDP-Cyber ($p \statPosPCyber{}$; \statPosPredACyber{} of predictions against
\statPosGoldACyber{} of correct answers) but not on WMDP-Bio ($p = \statPosPBio{}$; power \statPosPowerBio{} against the worst
case in which every error lands on the first option), and on Cyber it is confined to
uncertain items: at $\pmax <
0.5$, \statPosLowPredA{} of predictions fall first against \statPosLowGoldA{} of correct
answers ($p \statPosLowP{}$), whereas at $\pmax \geq 0.9$ it is absent ($p =
\statPosHighP{}$, power \statPosHighPower{}). Because the default prompt labels options in positional order, a preference for the
first \emph{position} and one for the letter \texttt{A} are confounded; two re-runs
separate them. Replacing the letters with non-alphabetic glyphs, which carry no ordinal
prior, leaves the effect intact, the first position taking \statAbSymCyber{} of WMDP-Cyber predictions against \statAbBaseCyber{} at baseline. Permuting the letters across
positions splits it, with \statAbShufPosCyber{} falling on the first position and
\statAbShufLetterCyber{} on the letter \texttt{A}. The preference is therefore primarily
positional, with a weaker label component. Quantisation ties do not explain it either: ties occur on only \statTieFrac{} of records
(\statTieTotal{} items, the first option among the tied set in
\statTieContainingFirst{}), and it wins \statTieObservedFirst{} of them against
\statTieExpectedFirst{} expected under random tie-breaking. Accuracy is barely affected by the labelling scheme (glyphs change it by
$\statAbSymAccDeltaCyber{}$~pp, 95\% CI $\statAbSymAccCICyber{}$), but prompt format
matters: listing the options only as category keys costs 4.7~pp on WMDP-Cyber (95\% CI
$\statAbCritCICyber{}$) and listing them only in the state text costs 2.0~pp
($\statAbStateCICyber{}$), so the redundant serialisation is doing work.

\subsection{Selective permutation averaging}
\label{sec:permavg}

If a single evaluation returns one of several possible answers, averaging over presentations
should help where instability is high. Averaging the four inverse-permuted probability distributions raises WMDP-Cyber accuracy
from \statSingleAccCyber{} (rotation~0 of the cyclic run; the independent single-pass
figure in Table~\ref{tab:main} is \statRCCyberBase{}) to \statEnsAccCyber{}
(\statEnsGainCyber{}~pp, 95\% CI $[\statEnsGainCyberLo{}, \statEnsGainCyberHi{}]$); majority
voting over the four answers is weaker (\statMajAccCyber{}). On WMDP-Bio, where
\statConsistBio{} of items are already stable, the gain is \statEnsGainBio{}~pp with an
interval spanning zero ($[\statEnsGainBioLo{}, \statEnsGainBioHi{}]$). Averaging helps on the
dataset that exhibits substantial instability and not on the one that does not.

Evaluating four rotations costs four times as much as one. Since $\pmax$ predicts
correctness, it can be used to select which items receive the additional evaluations.
Presenting a fraction $c$ of items under all four rotations, chosen least-confident first,
costs $1 + 3c$ calls per item. On WMDP-Cyber, selecting the least confident 20\% reaches
\statCascadeCyberAtTwenty{} and the least confident 40\% reaches \statCascadeCyberAtForty{}
at \statCascadeCyberCallsForty{} calls per item, which is \statCascadeFracOfGain{} of the
gain available from averaging every item at \statCascadeCostSaving{} lower cost
(Fig.~\ref{fig:permutation}(b)). Random selection at the same budget reaches
\statCascadeCyberRandForty{}, averaged over 400 draws, and no draw beat the confidence-based ordering. On WMDP-Bio the advantage over random selection is
\statCascadeAdvantageBio{}~pp and \statCascadeRandBeatBio{} of random draws do at least as
well, so we do not claim a benefit there. The distance to the oracle curve indicates the
headroom remaining for a better selection rule.

\section{Discussion and Limitations}

These results bear on how System-1 models should be evaluated. Aggregate accuracy was
nearly invariant across the four rotations while more than a third of WMDP-Cyber items
received different answers, so similar accuracy figures can conceal substantial item-level instability. Reported uncertainty is informative once its definition is
established: $\pmax$ separates correct from incorrect predictions with pooled AUROC
\statAurocPooled{} and supports abstention, but calibration quality and the error rate
at high confidence both vary by dataset, so a threshold chosen on one task should not be
assumed to hold on another. Stability under re-presentation is a second signal,
unavailable from a single evaluation, and repeated inference is worth its cost where
instability is high but not where it is low.

Where users can influence how a decision is presented, representation sensitivity could
become a manipulation surface, and disagreement across re-presentations could signal an
unreliable decision. Our benchmark does not test such a deployment.

These properties are relevant if System-1 models are later considered as components of
biosecurity classifiers or language-model safety pipelines. Evaluating such a use would
require a dedicated benign-versus-hazardous classification benchmark, which this study does
not provide.

\subsection{Limitations}

We evaluate one proprietary closed-weight model that exposes no seed, so these results
do not generalise to System-1 models as a class and exact reproduction depends on vendor
behaviour. We evaluate no biosafety or hazard-classification task, and multiple-choice
questions are only a proxy for the knowledge they probe; label quality in biology-adjacent sets can be poor~\cite{mmluredux} and we did not audit it, nor can benchmark
contamination be excluded. We did
not measure latency and did not run an external larger model as a fallback, so we make no
claim about either. Cyclic evaluation covers four rotations rather than all $4!$ permutations and only the
four-option WMDP suites, making the reported instability a lower bound. The label and
format ablations were run on those two suites only, so their conclusions do not
automatically extend to the LAB-Bench subtasks. Returned probabilities are quantised
to two decimals, and several LAB-Bench high-confidence subsets contain fewer than 25 items,
so we report intervals throughout.

\section{Conclusion}

We audited a commercial non-generative System-1 model on \statNItems{} items drawn from
biosecurity-relevant and biology-relevant benchmarks. Its accuracy is strongly
task-dependent, its top-1 probability is reasonably well calibrated and predicts its own
errors well enough to support abstention, and its decisions are considerably less stable
under reordering of the answer options than aggregate accuracy suggests. Averaging
probabilities over answer permutations improves accuracy where instability is high, and
applying it only to low-confidence items recovers most of that improvement at a fraction of
the cost. Confidence and stability under re-presentation provide complementary evidence
about when such a model's predictions should be trusted, and neither is visible in a single
accuracy figure.

\section*{Acknowledgment} This work has been supported by start-up funds awarded to
I.G.S. We thank TypeSafe for API access; the vendor had no role in the design, analysis,
or conclusions of this study and did not review this manuscript before submission.
\section*{Ethics and Conflict of Interest} This study authored no new hazardous
technical content: all items come from existing public benchmarks, and we release model
decision records but no new question content. We reported the \texttt{confidence} field
semantics to the vendor as a documentation issue. The authors declare no competing
financial or non-financial interests. \section*{Code Availability} \sloppy
Decision records, analysis code, and manuscript source are available at
\url{https://github.com/Georgakopoulos-Soares-lab/biosafety_knowledge_jev}.
Every reported statistic can be recomputed from them.

{\footnotesize
\bibliographystyle{IEEEtran}
\bibliography{refs}
}

\section*{LLM Usage Statement}
A large language model (LLM) based coding assistant (Anthropic Claude) was used to write
the evaluation harness
and analysis scripts, to assist with statistical analysis and literature search, and for
editorial work on this manuscript. All generated code was executed and its outputs
inspected by the authors. Every statistic in the running text is emitted by the released
pipeline, and the build fails if a cited value does not match it; all references were
verified against primary sources. An earlier version of this analysis contained statistical
errors, found and corrected during internal adversarial review. The model under evaluation
is a commercial closed-weight system, a reproducibility limitation we cannot remove. The
authors take full responsibility for the correctness, originality, and integrity of all
content.

\end{document}